\documentclass{sig-alternate}
\newfont{\mycrnotice}{ptmr8t at 7pt}
\newfont{\myconfname}{ptmri8t at 7pt}
\let\crnotice\mycrnotice%
\let\confname\myconfname%

\usepackage[german,american]{babel}
\usepackage[T1]{fontenc}
\usepackage[utf8]{inputenc}
\usepackage{times}
\usepackage{textcomp}
\usepackage{graphicx}
\usepackage{amssymb}
\usepackage{microtype} 

\usepackage{url}
\usepackage{enumitem}
\renewcommand{\tt}{\fontencoding{OT1}\fontfamily{cmtt}\selectfont}

\usepackage[usenames,dvipsnames]{color}
\usepackage{booktabs}
\usepackage[numbers,sectionbib]{natbib}
\makeatletter
\def\@copyrightspace{\relax}
\makeatother

\begin{document}

\title{Clickbait detection using word embeddings}
\subtitle{The torpedo Clickbait Detector at the Clickbait Challenge 2017}

\numberofauthors{2}
\author{
\alignauthor
Vijayasaradhi Indurthi\\
\affaddr{International Institute of Informatin Technology, Hyderabad}\\
\affaddr{vijaya.saradhi@research.iiit.ac.in}\\
\alignauthor
Subba Reddy Oota\\
\affaddr{International Institute of Information Technology, Hyderabad}\\
\affaddr{subbareddy.oota@students.iiit.ac.in}\\
\alignauthor
}

\maketitle

\begin{abstract}
Clickbait is a pejorative term describing web content that is aimed at generating online advertising revenue, especially at the expense of quality or accuracy, relying on sensationalist headlines or eye-catching thumbnail pictures to attract click-throughs and to encourage forwarding of the material over online social networks. 
We use distributed word representations of the words in the title as features to identify clickbaits in online news media. We train a machine learning model using linear regression to predict the cickbait score of a given tweet. Our methods achieve an F1-score of 64.98\% and an MSE of 0.0791. Compared to other methods, our method is simple, fast to train, does not require extensive feature engineering and yet moderately effective.
\end{abstract}

\section{Introduction}
Clickbait is that web content whose main purpose is to attract attention and encourage visitors to click on a link to a particular web page. Examples of such clickbaits include 
 
\begin{itemize}
\item \textit{``21 Completely Engrossing Fan Fictions You Won't Be Able To Stop Reading"}
\item \textit{``These White Tiger Cubs Are The Most Beautiful Creatures You'll See Today"} 
\item \textit{ ``Here's What Real Vegans Actually Eat"}
\item \textit{ ``Bow Wow Had No Clue How To Kill Time During The Grammys And It Was Hilarious"}
\item \textit{ ``We Know Who Your Celebrity Husband Should Be Based On One Question"}

\end{itemize}
Clickbaits employ the cognitive phenomenon known as Curiosity Gap \cite{Lowenstein:1994}, where the headlines provide forward referenced cues which generate sufficient curiosity compelling the reader to click the link and fill their curiosity gap.
Clickbaits eventually cause disappointment, as they are not able to live up to the promises made in the headline. Due to their heavy use in online journalism, it is important to develop techniques that automatically detect and combat clickbaits.

Research has shown that using distributed word embeddings can improve the performance of text classification as they capture lexical and semantic features of the text, without the need for explicit feature engineering. However, these word embeddings are generic and may not capture domain specific knowledge necessary for the classification task. Our motivation for this work is to specifically answer this question - "Can we use distributed word embeddings to train a machine learning model and predict the rating of a clickbait item". By conducting experiments on the clickbait dataset we show that this approach can improve the performance of the classification task.

The main contributions of our paper are as follows: (1) We identify a few hand-crafted features which capture domain specific information and use them for the classification task. We use pre trained GloVe vectors as features for the classification task (4) We augment the GloVe embeddings along with hand-crafted features for predicting the clickbait score of a tweet. Our methods achieve an F1-score of 64.98\% and an MSE of 0.0791.

\section{Related Work}
\cite{chakraborty2016stop} highlighted many interesting differences between clickbait and non-clickbait categories which include sentence structure, word patterns etc. They rely on a rich set of 14 hand-crafted features to detect clickbait headlines. In addition, \cite{chakraborty2016stop} build a browser extension which warns the readers of different media sites about the possibility of being baited by such headlines. Their methods achieve 93\% accuracy in detecting and 89\% accuracy in blocking clickbaits.

\cite{potthast2016clickbait} attempted to detect clickbaity Tweets in Twitter by using common words occurring in clickbaits, and by extracting some other tweet specific features. They achieve an F1 score of 73\% in classifying tweets as clickbaits or not.

\cite{chen2015misleading} argued for labeling clickbaits as misleading content or false news. 

\cite{anand2016we} used deep learning techniques like Bi-Directional Recurrent Neural Network model with character and word embeddings as the features. They achieve the state of the art results with an F1 score of 98\% in classifying online content as clickbaits or not. 

While \cite{chakraborty2016stop} and \cite{anand2016we} explore identifying clickbaity titles in webpages, \cite{potthast2016clickbait} explore identifying clickbaits in tweets.

Unlike earlier work done on clickbaits, the clickbait challenge \cite{potthast:2017a} requires us to calculate a clickbait score of a tweet post.

\section{Approach}
The clickbait dataset \cite{potthast:2017b} contained tweets from Twitter. Twitter is an online news and social networking service where users post and interact with messages, "tweets", restricted to 140 characters. Each tweet in the dataset has the text of the posted tweet and its associated metadata like keywords, time of the post, media linked with the post, description of the target and the target paragraphs. 

In spite of the availability of the tweets' metadata, we limit our experiments to only the text of the post for training a machine learning model to predict the clickbait score of each tweet. 

We augment a few hand-crafted domain specific features along with pre-trained distributed word representations as features for this task. We train a linear regression model to predict the clickbait score of a tweet.

\begin{table}
\centering
\begin{tabular}{l}  \hline \hline
{Features Used}\\ \hline \hline
Number of words \\ 
Number of stop words \\
Average length of the word  \\
Presence of question form \\
Presence of numbers at the start of headline \\
Presence of continuous form of verb \\
Presence of superlative forms of adjectives\\
300 dimensions of the GloVe embeddings\\
\hline \hline
\end{tabular} 
\caption{Features used for training our model.}
\label{comparison}
\end{table}

\textbf{Hand-crafted features:}
In addition to using the first three features used by \cite{chakraborty2016stop} i.e number of words, number of stopwords and the average word length of the clickbait headlines,  we attempt to use the following additional hand-crafted features.

\begin{enumerate}[topsep=5pt,itemsep=-1ex,partopsep=1ex,parsep=1ex]
\item Presence of question form - When, What, Which, Who, When, Whose, Whom, How, Where, Which, Can, Should
\item Presence of digits at the beginning of the headline
\item Presence of gerunds i.e continuous form of the verb in the headline like walking, eating, attending etc.
\item Presence of superlative forms of adjectives like cutest, best, hottest, greatest etc. 

\end{enumerate}
\textbf{Distributed word embeddings:}
Distributed word embeddings map words in a language to high dimensional real-valued vectors in order to capture hidden semantic and syntactic properties of words. These embeddings are typically learned from large unlabeled text corpora. In our work, we use the pre-trained
300 dimensional GloVe embeddings ~\cite{pennington2014glove} which were trained on about 6 billion words from the 2014 Wikipedia corpus and English Gigaword Fifth Edition corpus using the Continuous Bag of Words architecture.

To arrive at the embedding of a tweet post, we take the average of the GloVe embeddings of all the words present in the tweet post.

We used Linear Regression technique, which is a very simple machine learning learning algorithm for predicting the clickbait score of the tweet post. We model the given challenge as a regression problem, where the dependent variable is the clickbait score and the independent variables are the features mentioned above. We had 307 features for training(7 handrafted and 300 from GloVe). We have not used any kind of regularisation technique. We felt that a simple model like Linear Regression would generalize sufficiently well than a complex model.

The advantages of our approach are
\begin{enumerate}[topsep=5pt,itemsep=-1ex,partopsep=1ex,parsep=1ex]
\item Simple to implement as there is not much feature engineering
\item Pretrained vectors are available which are ready to use
\item Machine learning technique is simple to train and does not need long training times.
\item Unlike deep learning methods, our methods are interpretable to a certain extent 
\item Can work with modest hardware requirements as training is not memory intensive.
\item Model generated is compact and can be used in low cost hardware like phones etc.

\end{enumerate}

\section{Evaluation Results}
2 datasets have been provided for training the model for \cite{potthast:2017a}. A small initial dataset used in \cite{potthast:2016} for training and a bigger dataset \cite{potthast:2017b} has been provided for validation. \ref{dataset} shows the details of each dataset.
In our approach, we concatenate the two datasets, both the training and validation dataset to make a bigger training dataset. Out of this we select the same number of clickbaits and nonclickbaits to have equal representation of the classes. From this, we randomly split the set into 80:20 for training and validation. 

For the final official evaluation, we have used the whole of the above dataset for training the model, which was used to make predictions on the unseen test set.

Official evaluation has been done on the platform called TIRA \cite{potthast:2014}.

\begin{table}
\centering
\begin{tabular}{l|r|r|r}  \hline \hline
{Dataset}&Total&Clickbaits&No-Clickbaits\\ \hline \hline
\cite{potthast:2016}Training&2495&762&1697 \\ 
\cite{potthast:2017b}Validation&19538&4761&14777 \\ 
Unlabelled&80012&NA&NA \\
Test&Unknown&Unknown&Unknown \\
\hline \hline
\end{tabular} 
\caption{Dataset details.}
\label{dataset}
\end{table}

\begin{table}
\centering
\begin{tabular}{l|l}  \hline \hline
Evaluation Metric&Value\\ \hline \hline
Mean squared error&0.0791655793621 \\ 
Median absolute error&0.236312405103 \\ 
F1 score&0.649884407912 \\ 
Precision&0.5297319933 \\ 
Recall&0.840531561462 \\ 
Accuracy&0.784551346225 \\ 
Normalised mean squared error&1.07669865075 \\ 
Mean absolute error&0.240963463871 \\ 
Explained variance&0.345845579242	 \\ 
R2 score&-0.0766986507455 \\ 
Runtime&00:04:55 \\ 
\hline \hline
\end{tabular} 
\caption{Results of official Evaluation for our method}
\label{results}
\end{table}
\ref{results} shows various evaluation metrics which the official evaluators have evaluated for our model.

The MSE for the baseline system was 0.0435, which our system was unable to achieve. This might be because of high bias in our model as our model was a very simple model. Selecting a little complex model or a machine learning technique, or with more feature engineering might help in improving the performance of the model.

\section{Conclusion}

In this paper, we develop a machine learning model, using pretrained distributed representations of words trained on a huge corpus, to predict the clickbait score of a tweet post.
In future, we would like to use a more complex machine learning models like neural network to build a model to predict the clickbait score of the clickbaits. We would want to include more hand crafted features in our future work.

\begin{raggedright}
\bibliography{clickbait17-notebook-lit}
\end{raggedright}
\end{document}